\documentclass[a4paper, 10pt, conference]{ieeeconf}      
\usepackage{FG2025}
\usepackage{amsmath}
\usepackage{algorithm}
\usepackage{algpseudocode}
\usepackage{graphicx}
\usepackage{booktabs}
\usepackage{adjustbox}
\usepackage{multirow}
\usepackage{subfigure}
\usepackage{amssymb}
\usepackage{hyperref}

\FGfinalcopy 

\IEEEoverridecommandlockouts                              
\def\FGPaperID{0217} 

\title{\LARGE \bf
Improved Ear Verification with Vision Transformers and Overlapping Patches
}

\author{%
  \parbox{\textwidth}{\centering
    {\large Deeksha Arun$^{1}$, Kagan Ozturk$^{1}$, Kevin W. Bowyer$^{1}$, Patrick Flynn$^{1}$}\\
    {\normalsize
      $^{1}$Department of Computer Science and Engineering, University of Notre Dame, Notre Dame, IN 46556
    }\\
  }
}

\begin{document}

\ifFGfinal
\thispagestyle{empty}
\pagestyle{empty}
\else
\author{Anonymous FG2025 submission\\ Paper ID \FGPaperID \\}
\pagestyle{plain}
\fi
\maketitle

\begin{abstract}

Ear recognition has emerged as a promising biometric modality due to the relative stability in appearance during adulthood. Although Vision Transformers (ViTs) have been widely used in image recognition tasks, their efficiency in ear recognition has been hampered by a lack of attention to overlapping patches, which is crucial for capturing intricate ear features. In this study, we evaluate ViT-Tiny (ViT-T), ViT-Small (ViT-S), ViT-Base (ViT-B) and ViT-Large (ViT-L) configurations on a diverse set of datasets (OPIB, AWE, WPUT, and EarVN1.0), using an overlapping patch selection strategy. Results demonstrate the critical importance of overlapping patches, yielding superior performance in 44 of 48 experiments in a structured study. Moreover, upon comparing the results of the overlapping patches with the non-overlapping configurations, the increase is significant, reaching up to 10\% for the EarVN1.0 dataset. In terms of model performance, the ViT-T model consistently outperformed the ViT-S, ViT-B, and ViT-L models on the AWE, WPUT, and EarVN1.0 datasets. The highest scores were achieved in a configuration with a patch size of 28×28 and a stride of 14 pixels. This patch-stride configuration represents 25\% of the normalized image area (112x112 pixels) for the patch size and 12.5\% of the row or column size for the stride. This study confirms that transformer architectures with overlapping patch selection can serve as an efficient and high-performing option for ear-based biometric recognition tasks in verification scenarios.

\end{abstract}

\section{INTRODUCTION}

    Ear recognition as a biometric modality has a long history, dating back many decades~\cite{bertillon1896signaletic}, and has seen attention as a possible alternative to traditional face recognition approaches. Ears have a relatively stable structural shape throughout an individual's adult lifetime \cite{sforza2009age, yoga2017assessment} and their images may be captured from a distance with little or no subject involvement, making ear recognition an appealing alternative to other modalities. Early studies on automatic ear recognition began in the mid-1990s, with researchers using geometrical analysis of ear contours and structures \cite{bolle1998biometrics, moreno1999use}. More recent and advanced methods have displayed improved robustness to effects such as illumination and pose variations~\cite{galdamez2017brief}.
    
    The accuracy of ear recognition systems has improved significantly in the last decade \cite{oyebiyi2023systematic, benzaoui2023comprehensive}, primarily driven by the advancements in deep learning methods. The large Unconstrained Ear Recognition Challenge (UERC) datasets ~\cite{emervsivc2019unconstrained,emervsic2023unconstrained} have played a critical role in reigniting interest as they capture a variety of ear appearances encountered in real-world scenarios—such as variances in pose, illumination, occlusion, and resolution.

    \begin{figure}[th]
      \centering
      
      \includegraphics[width=1.0\linewidth]{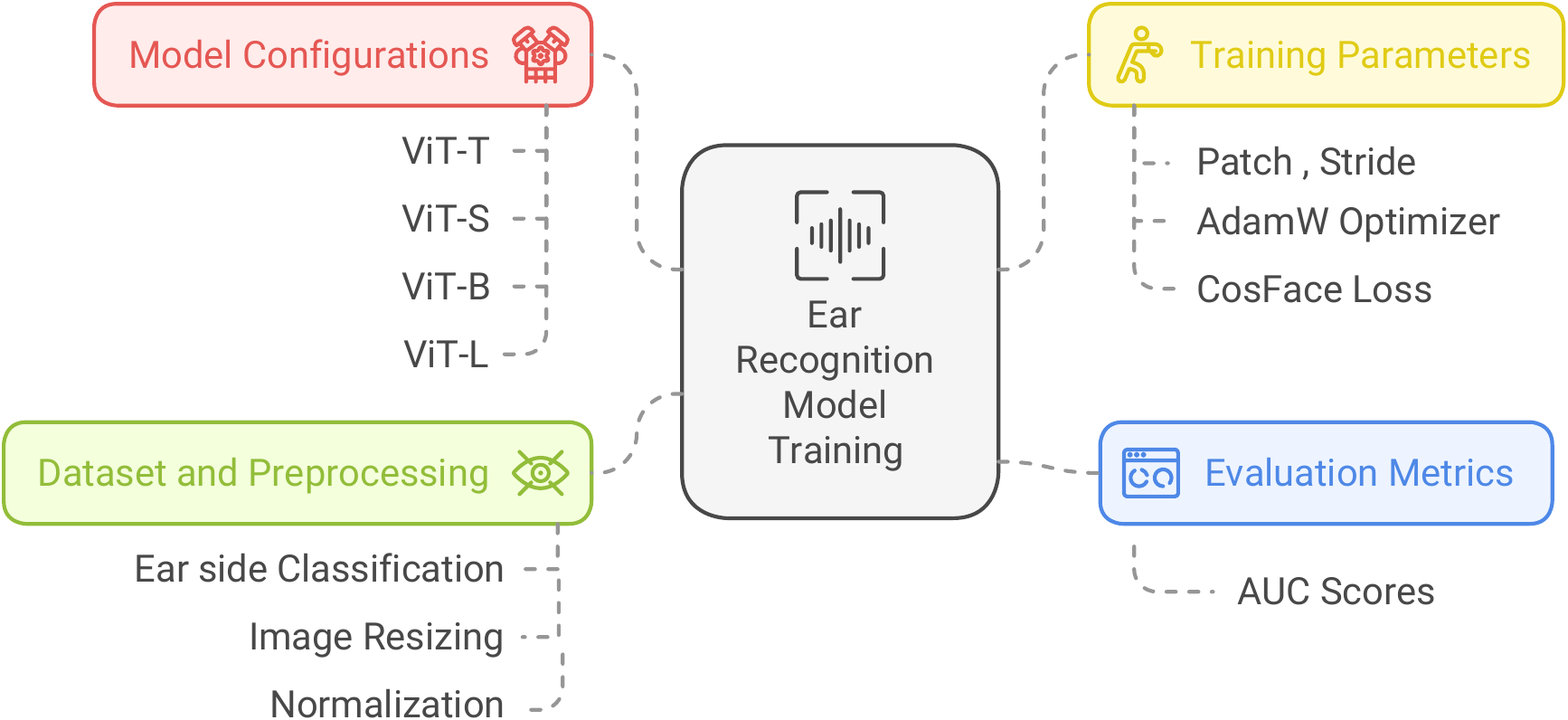}
       \caption{A schematic representation of the ear recognition model trained using Vision Transformers (ViTs)}
       \label{fig:ear_block_diagram}
       
    \end{figure}
\noindent 

Recent advances in computer vision include the emergence of Vision Transformers (ViTs)~\cite{dosovitskiy2020image} for recognition. These networks employ self-attention processes to identify both local and global patterns in image data. ViTs operate on a tokenized input consisting of image patches, allowing them to concentrate on broader contextual relationships than convolutional neural networks (CNNs), which are dependent on fixed receptive fields~\cite{ma2023visualizing}. This patch-based approach makes ViTs particularly flexible for various image recognition applications, including ear recognition. The ViTEar work ~\cite{emervsic2023unconstrained} fine-tuned pre-trained DINOv2 Vision Transformers ~\cite{oquab2023dinov2} with datasets like UERC2023 ~\cite{emervsic2023unconstrained} and EarVN1.0 \cite{hoang2019earvn1} to obtain a rank-1 accuracy of 96.27\% on the UERC2019 dataset ~\cite{emervsivc2019unconstrained}. Another research team~\cite{mehta2023vision} combined CNNs with ViTs to capitalize on their individual strengths, obtaining 99.36\% and 91.25\% accuracy on the Kaggle \cite{Kaggle_Ear_Dataset} and IITD-II \cite{IITD-II_Ear_Dataset} datasets, respectively. Furthermore, research ~\cite{alejo2021unconstrained} comparing ViT ~\cite{dosovitskiy2020image} and DeiT-based ~\cite{touvron2021training} models to CNNs found that transformer models outperformed CNNs even without extensive data augmentation.

\begin{figure*}[tbhp]
\centering
  \includegraphics[width=0.99\linewidth,clip=]{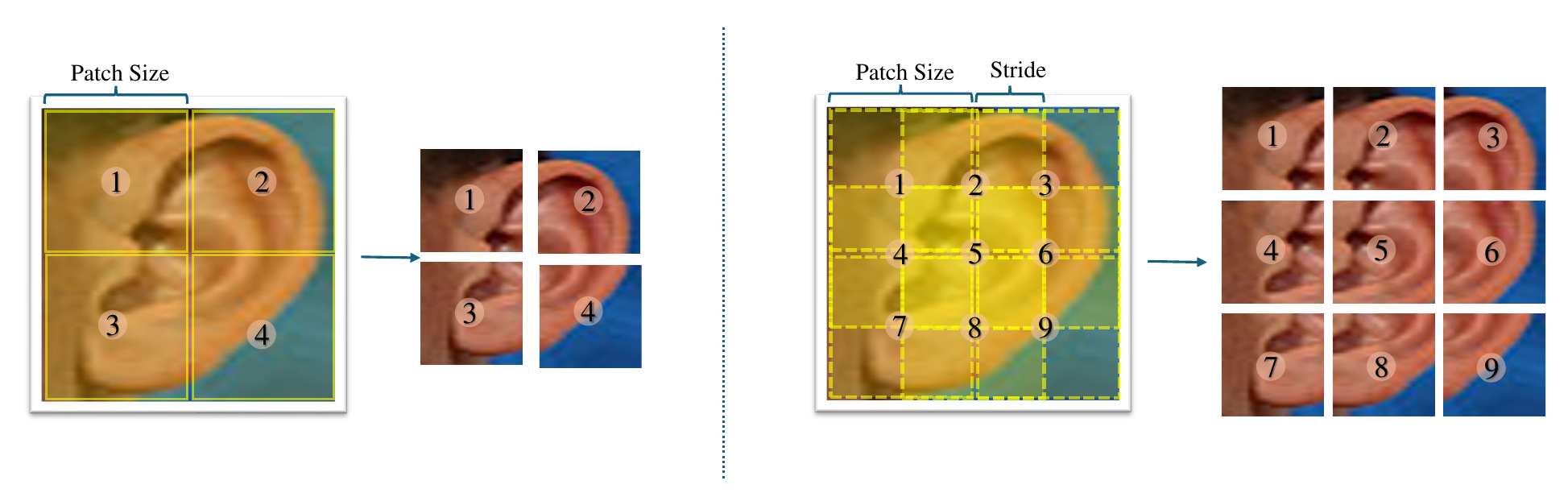}
  \caption{\textbf{Comparison of overlapping and non-overlapping patches:} In this example, we consider the cases- p56\_s56 (left part--non-overlapping patches) and p56\_s28 (right part--overlapping patches), where p56\_s56 implies a patch size of 56 and a stride of 56, and p56\_s28 represents a patch size of 56 and a stride of 28. These configurations result in 4 and 9 patches respectively, which can be calculated using~\eqref{NPatch}. The right part shows how the stride setting captures inter-patch information, thus aiding in the improvement of ear verification performance.}
  \label{fig:over}
\end{figure*}

Previous works exploring the use of ViTs for ear recognition~\cite{alejo2021unconstrained,mehta2023vision} trained ViTs on smaller datasets and evaluated one or two datasets with fewer subjects in the test set due to computational constraints.
%
In addition, the existing work did not explore the use of overlapping patches~\cite{zhong2021face}, which can more effectively model features that extend beyond a patch boundary, leading to improved feature representation. Overlapping patches help capture both the small details and the larger, continuous structures of the human ear. Figure~\ref{fig:over} illustrates the comparison between overlapping and non-overlapping patches.

%

In our work, as shown schematically in Fig.~\ref{fig:ear_block_diagram}, we train customized ViT models (obtained from the Insightface repository) \cite{an2022partialfc} using T, S, B, and L  configurations on the UERC2023 dataset ~\cite{emervsic2023unconstrained} (excluding the AWE component) and evaluate it on four datasets: OPIB \cite{Adebayo2023}, AWE \cite{emervsivc2017ear}, WPUT \cite{frejlichowski2010west}, and EarVN1.0 \cite{hoang2019earvn1}. We also explore the impact of overlapping patches in ViTs. By employing a large dataset for training, we can reduce the risk of over-fitting during training. Furthermore, testing on several different datasets gives a broader understanding of how the trained model reacts to variations in data distribution and subject demographics, revealing the model's generalizability to multiple settings.

The contributions of this work are:
\begin{enumerate}
\item A systematic study of the use of four transformer models trained on the same large and diverse data set.

\item The results of experiments with several patch-stride configurations, providing a comprehensive evaluation of the impact of different patch size and overlapping stride settings.

\end{enumerate}

Our work establishes a robust benchmark for ear recognition in a verification scenario using ViTs, paving the way for future research and advancements in this domain.

\begin{table*}[!ht]
\caption{\MakeUppercase{Comparison of Test Datasets}}
\renewcommand{\arraystretch}{1.5} 
\centering
\begin{tabular}{|p{3cm}|p{2.5cm}|p{2.5cm}|p{2.5cm}|p{2.5cm}|}
\hline
\textbf{Characteristics} & \textbf{OPIB} & \textbf{AWE} & \textbf{WPUT} & \textbf{EarVN1.0} \\
\hline
Images & 907 & 1000 & 2071 & 28,412 \\
\hline
Subjects  & 152 & 100 & 501 & 164 \\
\hline
Images per Subject & 6 & 10 & $\sim$4 & 100+ \\
\hline
Gender Balance & 59.65\% M, 40.35\% F & 91\% M, 9\% F & 50.70\% F, 49.30\% M & 59.76\% M, 40.24\% F \\
\hline
Ethnicity & African & Diverse (61\% white, 18\% Asian, 11\% black, etc.) & Diverse (White (W), Black (B), Yellow (A), Other (O)) & Asian \\
\hline
Occlusion & Present & Present & Present & Present \\
\hline
Pose Information & Specific pose angles & Dynamic poses & Specific pose angles & Dynamic poses \\
\hline
Head Side & Specified (L and R) & Specified (L and R) & Specified (L and R) & Not specified \\
\hline
Accessories & Headphones, scarves, earrings & Earrings, other accessories & Earrings, spectacles, other accessories & Large ear accessories and mixed with other non-interest regions including hair and face skin\\
\hline
\end{tabular}

\label{tab:ear_datasets}
\end{table*}

\section{RELATED WORK}

The human ear's complex anatomy and development have been extensively studied in medical and anatomical literature \cite{alvord1997anatomy, mozaffari2021anatomy}. Its structure, including the helix, antihelix, concha, tragus, and other components, plays a crucial role in both its auditory function and its distinctiveness for recognition applications \cite{benzaoui2023comprehensive}. Ears have a relatively stable structural shape throughout an individual's adult lifetime~\cite{sforza2009age, yoga2017assessment}, making them suitable for recognition. Traditional ear detection and recognition methods rely on the ear's physiological features, such as its shape, contours, and geometric structure~\cite{dodge2018unconstrained, alshazly2019handcrafted, moreno1999use, choras2010ear}. Challenges such as occlusion or changes in pose and lighting, which can distort the ear’s features, can often limit the performance of traditional methods~\cite{alva2019review}. Subspace learning techniques like PCA, LDA, and ICA have been effective for local ear contour feature extraction \cite{alva2019review}. Hybrid methods combining multiple attributes and dimensionality reduction offer higher recognition performance but are more computationally expensive~\cite{alva2019review, omara2016novel, alshazly2019handcrafted}. All of these traditional ear recognition methods face significant challenges, particularly in unconstrained, "in the wild" conditions. The emergence of deep learning led to new attention to the ear recognition problem~\cite{ying2018human, abdellatef2020fusion}, in the hopes of improving accuracy and robustness.

The first major application of convolutional neural networks (CNNs) for ear recognition was introduced by Galdámez {\em et al.}~\cite{galdamez2017brief} in 2017. Their study compared CNN-based methods with traditional approaches like LDA, PCA, and SURF, demonstrating that CNNs were more robust and effective than the traditional methods. Omara {\em et al.}~\cite{omara2018learning} in 2018 proposed a novel approach utilizing deep hierarchical features, where a CNN pre-trained on a large dataset was used to extract deep features. These features were then aggregated through deep characteristic aggregation (DCA) to improve feature representation and reduce dimensionality. They solved the resulting binary classification problem using an SVM classifier. Dodge {\em et al.}~\cite{dodge2018unconstrained} in 2018 employed transfer learning with deep neural networks (DNNs) for ear identification, utilizing data augmentation techniques and shallow classifiers. They recommended using deep averaging to reduce overfitting when working with smaller datasets.

In 2018, Zhang {\em et al.}~\cite{zhang2018ear} presented the Helloear dataset, which contains more than 600,000 images, and performed verification tests by modifying CNN architectures. They applied spatial pyramid pooling to handle images of varying sizes and used center loss to create more robust features, proving the effectiveness of multi-level CNN concatenation.  The Helloear dataset does not appear to be available at the time this paper was written. Eyiokur {\em et al.}~\cite{eyiokur2018domain} emphasized the importance of domain adaptation, showing that fine-tuning models on ear-specific datasets resulted in better performance than using generic datasets. In another study, Alshazly {\em et al.}~\cite{alshazly2019handcrafted} compared CNNs with local-texture descriptors, finding that CNN-based models outperformed traditional methods. They later enhanced their approach by using an ensemble of deep VGG networks for improved performance. In 2020, Alshazly {\em et al.}~\cite{alshazly2020deep} compared several deep CNN architectures and domain adaptation strategies on the EarVN1.0 database, concluding that fine-tuned ResNeXt101 models improved the recognition accuracy.
 
Recent studies have tackled the challenge of limited training data in CNNs for ear recognition. Emer\v{s}i\v{c} {\em et al.}~\cite{emervsivc2017training} employed data augmentation techniques such as rotation, scaling, and transformations to expand their dataset and enhance recognition. Zhang {\em et al.}~\cite{zhang2019few} in 2019 combined few-shot learning with data augmentation to address the issue of data scarcity. Transfer learning has also been widely used to fine-tune pre-trained models for improved domain-specific performance~\cite{tan2018survey}. 

In the past few years, the field of ear recognition has advanced through various approaches, with different directions explored to enhance its applicability. Khaldi and Benzaoui~\cite{khaldi2021new} recently developed a method to restore color information in grayscale images using deep conditional generative adversarial networks (DCGANs), enhancing recognition accuracy by improving image quality. Khaldi {\em et al.}~\cite{khaldi2021ear} introduced deep unsupervised active learning (DUAL), allowing the model to improve after the initial training phase by utilizing test images in an unsupervised learning process. Recently, Priyadharshini {\em et al.}~\cite{ahila2021deep} examined CNN performance for ear identification by adjusting various parameters and showed its success in large-scale monitoring systems. 
Working in the same direction, Alshazly {\em et al.}~\cite{alshazly2021towards} and Korchi {\em et al.}~\cite{korichi2022tr} advanced ear recognition through the use of various CNN architectures, including ResNet versions, and unsupervised learning methods like tied-rank independent component analysis to improve feature normalization.

More recently, advanced approaches in ear recognition include transformer-based models, which have demonstrated promising results~\cite{alejo2021unconstrained,mehta2023vision,emervsic2023unconstrained}. The ViTEar method~\cite{emervsic2023unconstrained} achieved 96.27\% rank-1 accuracy on the UERC2019~\cite{emervsivc2019unconstrained} test dataset by using a pre-trained DINOv2 Vision Transformer~\cite{oquab2023dinov2} fine-tuned on the UERC2023~\cite{emervsic2023unconstrained} and EarVN1.0~\cite{hoang2019earvn1} datasets. Another work~\cite{mehta2023vision} uses ViTs and CNNs to achieve 99.36\% and 91.25\% accuracy on the Kaggle \cite{Kaggle_Ear_Dataset} and IITD-II \cite{IITD-II_Ear_Dataset} datasets, respectively. A third study~\cite{alejo2021unconstrained} uses ViT~\cite{dosovitskiy2020image} and Data-efficient image Transformers (DeiTs)~\cite{touvron2021training} for unconstrained ear detection, yielding accuracy equivalent to or better than cutting-edge CNN-based approaches without requiring considerable data augmentation. However, these models~\cite{alejo2021unconstrained, mehta2023vision} were trained on smaller datasets and did not make use of overlapping patches~\cite{zhong2021face}, which are critical for capturing fine anatomical details of the ear that often span across multiple patches and contribute significantly to the ear's biometric uniqueness. The absence of this capability potentially limits the models' effectiveness in recognition performance. In our work, we train ViT models (T, S, B, L configurations) using the vast and diverse UERC2023~\cite{emervsic2023unconstrained} dataset, excluding the AWE subsets \cite{emervsivc2017ear}, and use the recent improvement of overlapping image patches in Vision Transformers to capture inter-patch details, which is essential to improving ear verification accuracy, while comparing it to ViT models trained using non-overlapping patches across several diverse datasets, namely, OPIB~\cite{Adebayo2023} , AWE \cite{emervsivc2017ear}, WPUT \cite{frejlichowski2010west}, and EarVN1.0 \cite{hoang2019earvn1}.

\begin{figure*}[t]
\centering
  \includegraphics[width=0.99\linewidth,clip=]{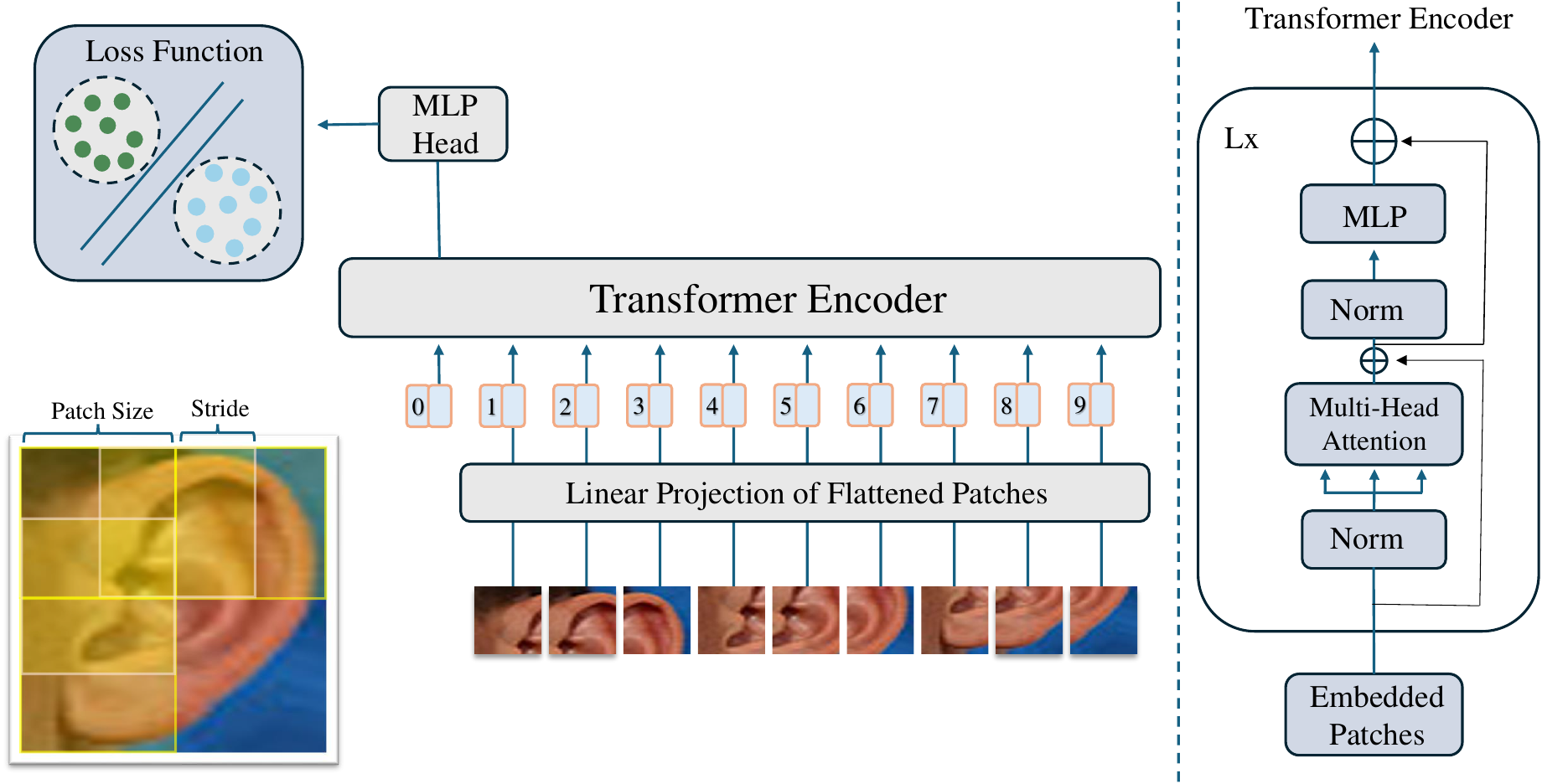}
  \caption{\textbf{Overview of the Ear Transformer}: The Ear Transformer splits images into overlapping patches, processed as tokens by the transformer encoder. This modification to ViT~\cite{dosovitskiy2020image} improves inter-patch relationships and performance. }
  \label{fig:overview}
\end{figure*}

\section{DATASET}

\subsection{Training data}

The UERC2023 \cite{emervsic2023unconstrained} dataset was used to train the vision transformers. It is a composite data set assembled from the UERC2017 and UERC2019 competition data sets (14,004 images from 650 subjects) and the VGGFace-Ear~\cite{ramos2022vggface} dataset (234,651 images from 660 subjects).  The VGGFace-Ear dataset was created by cropping ear regions from face images in the VGGFace~\cite{cao2018vggface2} dataset and resizing them to a fixed size. We excluded images from the first 100 subjects from the UERC2023 dataset that also belonged to the AWE~\cite{emervsivc2017ear} dataset (used below for testing), so the final number of subjects was 2404, and the total number of images was 247,655. UERC2023 dataset provides a wide variety of ear images encountered in real-world scenarios with variances in gender, ethnicity, poses, illumination and occlusion, making it suitable for model training.

\subsection{Testing data}

We used several distinct datasets for testing. The datasets used for testing include- OPIB, AWE, WPUT, and EarVN1.0, with each dataset considered independently as a separate test set. Table~\ref{tab:ear_datasets} provides a comparison of characteristics across all the test datasets.

\begin{enumerate}

\item The OPIB dataset~\cite{Adebayo2023} contains 907 images from 152 persons of African ethnicity. Three images of the left ear and three of the right ear were captured from most subjects at 0, 30, and 60-degree angles relative to a profile view looking directly at the ear. The dataset contains ear images with occlusions, like headphones, scarves, and earrings. Both indoor and outdoor settings were used to gather the photos. Students from a Nigerian public university collected the data; the gender distribution was 59.65\% male and 40.35\% female. 

\item AWE (Annotated Web Ears)~\cite{emervsivc2017ear}  contains 1000 images – exactly 10 images of each of 100 subjects. These images are close crops of the ears of famous persons and were collected through Web queries, which were then manually filtered. The gender breakdown is 91\% male and 9\% female. The ethnic composition is broad, with 61\% white, 18\% Asian, 11\% black, 3\% South Asian, 3\% Middle Eastern, 3\% South American, and 1\% other. For accessories, 91\% of the photographs show no accessories, 8\% show earrings, and 1\% show other accessories. Occlusion levels are classified as 65\% no occlusion, 28\% mild occlusion, and 7\% severe occlusion. The dataset also contains precise head position data with different distributions for pitch, roll and yaw. Furthermore, head side information is accessible, with 52\% labeled as Left and 48\% as Right. The dataset, therefore, includes a diverse collection of ear images with various attributes.

\item The West Pomeranian University of Technology (WPUT) dataset~\cite{frejlichowski2010west} contains 2071 cropped color images of the ears of 501 individuals. This dataset is rather diverse in environmental/imaging conditions, which were carefully captured and formed the basis for image file naming. Gender balance was approximately equal, with 50.70\% female and 49.30\% male. Images were taken indoors at 15.6\%, outdoors, and at 2\% dark. Two or more images of each of the left and right ears of each subject are provided in the dataset, with two choices of head pose (profile and 15 degrees frontal from profile). The distribution of ages was “0” to “above 50”, with the modal age being between 21 and 35 years of age. The dataset is dominated by subjects who appeared in only one imaging session. A significant proportion of the images (80\%) contain auricle deformations such as hair occlusions (166 subjects), earrings (147 subjects), spectacles, and other accessories, with the remaining 20\% devoid of occlusions. Motion blur artifacts were detected in 8\% of the photos, highlighting the difficulties in identifying the human ear under varying settings.

\item EarVN1.0~\cite{hoang2019earvn1} contains  28,412 unconstrained, low-resolution color ear images of 98 Asian men and 66 Asian women. The original facial images have been obtained in unconstrained conditions such as illumination, occlusion, and rotations, including camera systems and lighting conditions. Ear images are then cropped from facial images, featuring a variety of ear poses and illumination levels. Each subject is represented by 100 or more images in the dataset.

\end{enumerate}

\section{METHODOLOGY}

In this study, the customized ViT configurations used in~\cite{an2022partialfc}, namely  ViT-T, ViT-S, ViT-B, and ViT-L were used to train the ear recognition model in a verification scenario. These configurations differ in model size and complexity, thus enabling us to observe the effect of these configurations on the ear verification performance. The overview of the Ear Transformer used in this paper is shown in Figure~\ref{fig:overview}.

The ViT models were trained using the UERC2023 dataset, excluding the AWE subset, with standard pre-processing steps including image resizing ($W \times W$, with $W=112$) and normalization applied to the images. These images were divided into patches of size $P\times P$ with a stride of $S\times S$ (with both $P$ and $S$ being factors of $W$ to avoid truncated patches), yielding 
\begin{equation}
    N(W, P, S) = \left( \frac{W - P}{S} + 1 \right)^2
    \label{NPatch}
\end{equation}
resulting patches.

To understand the change when half stride size is used, we define the percentage variation ($PV$) to compare the performance of two patch-stride settings $A$ and $B$ as

\begin{equation}
    PV(A,B):= \left( \dfrac{AUC(A)-AUC(B)}{AUC(B)} \right) \times 100
    \label{eq:PV},
\end{equation}
where $AUC(\cdot)$ denotes the Area Under the Curve for the patch-stride setting $(\cdot)$.

%
%

Ensuring that the learned embeddings are highly discriminative and well-separated across classes is essential for achieving robust and reliable ear recognition, especially when dealing with variations in ear structure, orientation, and lighting conditions. An effective loss function plays a pivotal role in guiding the model's training process, enabling it to achieve high accuracy and strong generalization. Following the approaches proposed in~\cite{wang2018cosface}\cite{zhong2021face}, we employ softmax-based loss functions to ensure that the Vision Transformer effectively captures the refined features of ear images, leading to highly discriminative embeddings and superior performance in challenging scenarios.

To investigate the influence of spatial granularity on verification performance, patch sizes $P$ of 56, 28, and 16 were utilized. Stride values $S$ were chosen to arrange for either no overlap between patches or an overlap of one-half the patch extent (vertical and/or horizontal) to explore the trade-off between redundancy in token count/contents and performance.  The training was conducted using AdamW optimization with a learning rate of 0.001, a weight decay of 0.1, and a sample rate of 0.3. The models were trained over 100 epochs, including 10 warmup epochs, and mixed precision (fp16) was employed to enhance training efficiency.

\renewcommand{\arraystretch}{1.5} 
\begin{table*}[ht]
\centering
\caption{\ Comparison of AUC Scores across all ViT configurations including all patch-stride settings for OPIB, AWE, WPUT and EarVN1.0 datasets. Bold text highlights the performance improvements achieved with the use of overlapping patches.}

\begin{tabular}{|l|c|c|c|c|}
\hline
\textbf{Model Name} & \textbf{OPIB} & \textbf{AWE}  & \textbf{WPUT} & \textbf{EarVN1.0} \\
\hline

ViT\_T\_p16\_s8\     &         0.9078 $\pm$ 0.0042           &         0.9818 $\pm$ 0.0019         &         0.9396 $\pm$ 0.0032      & \textbf{0.7736 $\pm$ 0.0030}        \\
ViT\_T\_p16\_s16\    &         0.9168 $\pm$ 0.0061           &         0.9832 $\pm$ 0.0013         &         0.9442 $\pm$ 0.0055      &         0.7726 $\pm$ 0.0029         \\
ViT\_T\_p28\_s14\    & \textbf{0.9132 $\pm$ 0.0015}          & \textbf{0.9834 $\pm$ 0.0011}        & \textbf{0.9472 $\pm$ 0.0022}     & \textbf{0.7838 $\pm$ 0.0019}        \\
ViT\_T\_p28\_s28\    &         0.8926 $\pm$ 0.0055           &         0.9732 $\pm$ 0.0015         &         0.9286 $\pm$ 0.0047      &         0.7356 $\pm$ 0.0017         \\
ViT\_T\_p56\_s28\    & \textbf{0.8616 $\pm$ 0.0063}          & \textbf{0.9622 $\pm$ 0.0018}        & \textbf{0.9126 $\pm$ 0.0027}     & \textbf{0.6966 $\pm$ 0.0021}        \\
ViT\_T\_p56\_s56\    &         0.8258 $\pm$ 0.0065           &         0.9250 $\pm$ 0.0019         &         0.8782 $\pm$ 0.0082      &         0.6330 $\pm$ 0.0033         \\
                                                                                                                                  
\hline                                                                                                                            
                                                                                                                                  
ViT\_S\_p16\_s8\     &         0.9074 $\pm$ 0.0033           & \textbf{0.9800 $\pm$ 0.0012}         & \textbf{0.9390 $\pm$ 0.0032}    & \textbf{0.7644 $\pm$ 0.0029}        \\
ViT\_S\_p16\_s16\    &         0.9148 $\pm$ 0.0019           &         0.9778 $\pm$ 0.0013          &         0.9382 $\pm$ 0.0024     &         0.7536 $\pm$ 0.0017         \\
ViT\_S\_p28\_s14\    & \textbf{0.9090 $\pm$ 0.0037}          & \textbf{0.9800 $\pm$ 0.0007}         & \textbf{0.9418 $\pm$ 0.0030}    & \textbf{0.7700 $\pm$ 0.0022}        \\
ViT\_S\_p28\_s28\    &         0.8934 $\pm$ 0.0076           &         0.9640 $\pm$ 0.0019          &         0.9270 $\pm$ 0.0017     &         0.7164 $\pm$ 0.0017         \\
ViT\_S\_p56\_s28\    & \textbf{0.8696 $\pm$ 0.0078}          & \textbf{0.9528 $\pm$ 0.0019}         & \textbf{0.9058 $\pm$ 0.0059}    & \textbf{0.6818 $\pm$ 0.0018}        \\
ViT\_S\_p56\_s56\    &         0.8432 $\pm$ 0.0064           &         0.9132 $\pm$ 0.0041          &         0.8864 $\pm$ 0.0040     &         0.6296 $\pm$ 0.0026         \\
                                                                                                                                  
\hline                                                                                                                            
                                                                                                                                  
ViT\_B\_p16\_s8\     & \textbf{0.9022 $\pm$ 0.0143}          & \textbf{0.9216 $\pm$ 0.0088}         & \textbf{0.9432 $\pm$ 0.0029}    & \textbf{0.7216 $\pm$ 0.0055}        \\
ViT\_B\_p16\_s16\    &         0.8808 $\pm$ 0.0144           &         0.8828 $\pm$ 0.0194          &         0.9244 $\pm$ 0.0111     &         0.7086 $\pm$ 0.0118         \\
ViT\_B\_p28\_s14\    & \textbf{0.9100 $\pm$ 0.0041}          & \textbf{0.9238 $\pm$ 0.0075}         & \textbf{0.9452 $\pm$ 0.0033}    & \textbf{0.7324 $\pm$ 0.0047}        \\
ViT\_B\_p28\_s28\    &         0.9036 $\pm$ 0.0040           &         0.9044 $\pm$ 0.0032          &         0.9320 $\pm$ 0.0019     &         0.7278 $\pm$ 0.0033         \\
ViT\_B\_p56\_s28\    & \textbf{0.9196 $\pm$ 0.0026}          & \textbf{0.9208 $\pm$ 0.0016}         & \textbf{0.9444 $\pm$ 0.0017}    & \textbf{0.7364 $\pm$ 0.0029}        \\
ViT\_B\_p56\_s56\    &         0.8652 $\pm$ 0.0158           &         0.8796 $\pm$ 0.0072          &         0.9144 $\pm$ 0.0053     &         0.6978 $\pm$ 0.0065         \\
                                                                                                                                  
\hline                                                                                                                            
                                                                                                                                  
ViT\_L\_p16\_s8\     & \textbf{0.9190 $\pm$ 0.0040}          & \textbf{0.9816 $\pm$ 0.0021}         & \textbf{0.9458 $\pm$ 0.0022}    & \textbf{0.7774 $\pm$ 0.0044}        \\
ViT\_L\_p16\_s16\    &         0.9184 $\pm$ 0.0054           &         0.9712 $\pm$ 0.0018          &         0.9364 $\pm$ 0.0043     &         0.7358 $\pm$ 0.0018         \\
ViT\_L\_p28\_s14\    & \textbf{0.9190 $\pm$ 0.0058}          & \textbf{0.9750 $\pm$ 0.0016}         & \textbf{0.9400 $\pm$ 0.0031}    & \textbf{0.7536 $\pm$ 0.0057}        \\
ViT\_L\_p28\_s28\    &         0.9028 $\pm$ 0.0042           &         0.9610 $\pm$ 0.0037          &         0.9272 $\pm$ 0.0048     &         0.7176 $\pm$ 0.0046         \\
ViT\_L\_p56\_s28\    & \textbf{0.8822 $\pm$ 0.0041}          & \textbf{0.9514 $\pm$ 0.0044}         & \textbf{0.9158 $\pm$ 0.0028}    & \textbf{0.6966 $\pm$ 0.0038}        \\
ViT\_L\_p56\_s56\    &         0.8574 $\pm$ 0.0092           &         0.9228 $\pm$ 0.0020          &         0.9106 $\pm$ 0.0027     &         0.6606 $\pm$ 0.0030         \\

\hline
\end{tabular}%

\label{tab:model_metrics}
\end{table*}

\section{Experimental Setup}

The ear recognition models were trained in a verification scenario using four different configurations of Vision Transformers (ViTs)—L (Large), S (Small), B (Base), and T (Tiny)—on the UERC2023 dataset (excluding the first 100 subjects that belong to the AWE dataset). In this study, the dataset was initially pre-processed using a ResNet-100 based ear-side classifier to distinguish between the left and right ear, treating each ear side as a separate entity. As a result, each subject had two separate folders: one for their right ear and another for their left ear, with each side treated as an independent subject.

Six distinct models were trained for each ViT configuration considering patch sizes 16, 28 and 56. Three models used non-overlapping patches, with strides equal to the patch size and the other three models used overlapping patches, with a half patch size overlap. This approach allowed for a thorough examination of the impact of patch size and overlap on model performance. To assess the performance of the trained models, ear matching experiments were conducted on four separate test datasets: AWE, OPIB, EarVN1.0, and WPUT. For each experiment, the entire dataset was used for testing, which differs from prior research, where datasets were split into training and testing sets. Therefore, direct comparison with previous work is not feasible.

 Ear matching was performed by computing the dot product similarity scores of the 512 dimensional feature embeddings extracted from the ViT models. Genuine and impostor pairs were identified on the basis of the subject IDs. The model’s performance was assessed using the AUC score, which reflects its ability to differentiate between genuine and impostor pairs by analyzing the ROC curve, showing the trade-off between True Positive Rate (TPR) and False Positive Rate (FPR). The matching experiment was repeated five times to ensure the reliability of the results. The average Area Under the Curve (AUC) scores, along with empirical confidence intervals for each configuration and patch-stride combination, were computed and are presented in Table~\ref{tab:model_metrics}.

\section{RESULTS}
\subsection{Model based Performance Results}

In this study, we evaluated the performance of different Vision Transformer (ViT) models, including ViT-Tiny (ViT-T), ViT-Small (ViT-S), ViT-Base (ViT-B) and ViT-Large (ViT-L) with different patch and stride configurations across various datasets:  OPIB, AWE, WPUT and EarVN1.0. Table~\ref{tab:model_metrics} shows that the ViT-T model outperformed all the other models across most of the datasets. Specifically, the ViT-T model with $P=28$ and $S=14$  (ViT\_T\_p28\_s14) achieved the highest accuracy on AWE, WPUT and EarVN1.0 datasets with AUC values of 0.9834 ± 0.0011, 0.9472 ± 0.0022 and 0.7838 ± 0.0019, respectively.

Among the ViT-L models, ViT\_L\_p16\_s8\ achieved the best performance on OPIB (0.9190 ± 0.0040), AWE (0.9816 ± 0.0021), WPUT (0.9458 ± 0.0022) and EarVN1.0 (0.7774 ± 0.0044) datasets. ViT-S models, though competitive, did not surpass ViT-T and ViT-L in most of the cases, but ViT\_S\_p28\_s14\ showed promising results with 0.9800 ± 0.0007 on AWE, 0.9418 ± 0.0030 on WPUT and 0.7700 ± 0.0022 on EarVN1.0 datasets. ViT-B demonstrated relatively lower accuracy in many cases compared to ViT-T, ViT-L, and ViT-S, but it achieved impressive results on the OPIB dataset with a performance of 0.9196 ± 0.0026.

In conclusion, ViT-T configurations demonstrated the best overall performance across datasets, followed by ViT-L and ViT-S models, while ViT-B models performed relatively weaker.

\begin{figure*}[t]
\centering
  \includegraphics[width=0.99\linewidth,clip=]{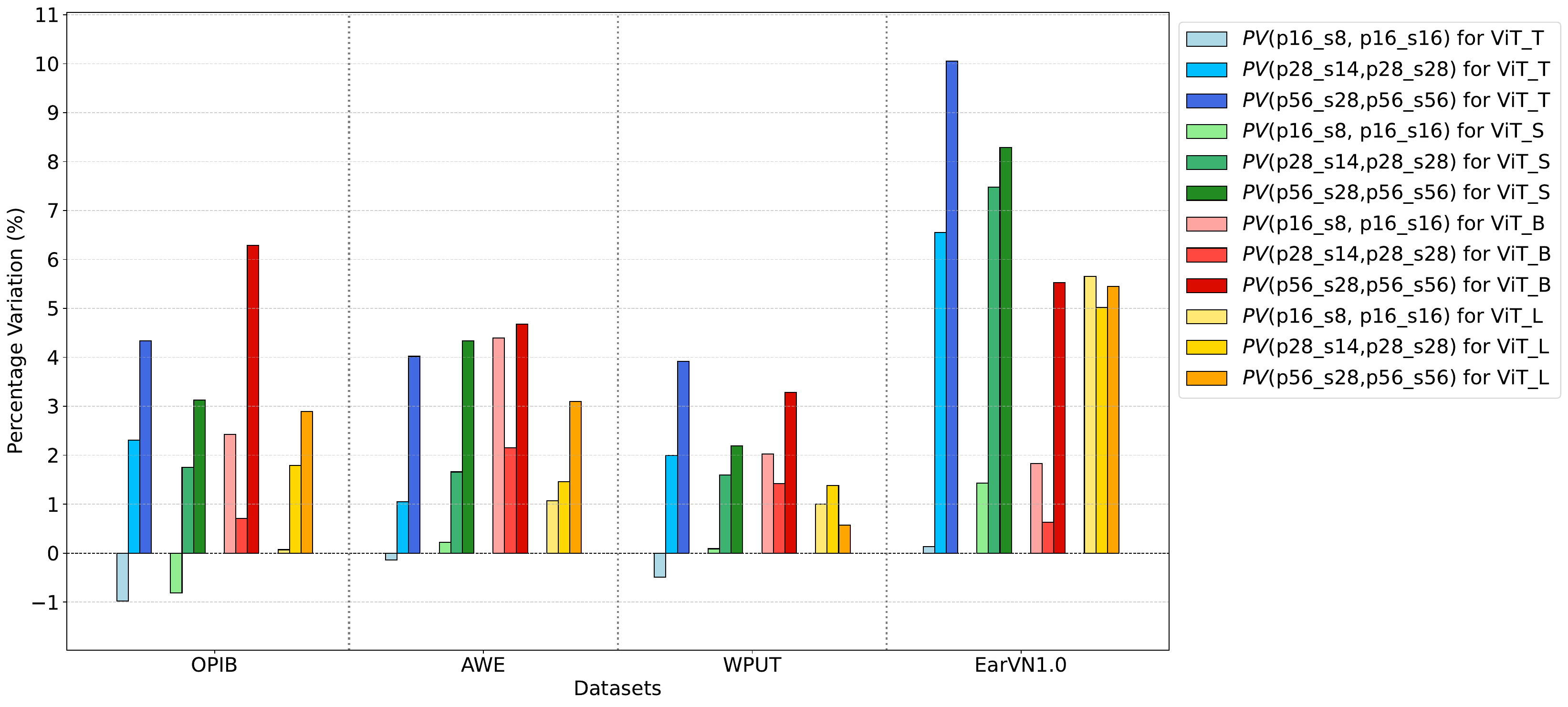}
  \caption{Percentage variation, $PV(A,B)$ as defined by~\eqref{eq:PV}, for all the datasets and for all the patch-stride pairs across all ViT Configurations (T, S, B and L). The setting $(A,B)$ refers to: $A:$ stride size is half of the patch size; $B:$ stride size is equal to the patch size.}
  \label{fig:PV}
\end{figure*}

\subsection{Patch-Stride based Performance Results}

Our experiments highlight the significant impact of stride on the ear verification performance (the stride size is set to half the patch size in our experiments). This effect is more prominent with larger patch sizes, such as $P=28$ and $P=56$. For instance, as shown in Table-\ref{tab:model_metrics}, configurations with $P=28$ (with a corresponding stride of $S=14$) or the configuration with $P=56, S=28$ consistently outperform the cases where $S=P$. This trend remains consistent across all configurations of the Vision Transformer (ViT) models analyzed in this study. The observed improvement suggests that larger patch sizes benefit significantly from a half-size stride.

However, when a smaller patch size was used ($S=16, P=8$), the larger models (ViT-L, ViT-B) consistently demonstrated enhanced performance with the half-patch stride, improving accuracy by 1-5\%. In contrast, smaller models (ViT-T, ViT-S) showed mixed results, with at most a marginal 1\% performance dip in only 4 out of 48 cases. To witness the increase and dip across all patch-stride pairs more clearly, we plot the bar graph of percentage variation, $PV (A, B)$ as defined by~\eqref{eq:PV}, for all the datasets across all ViT configurations (T, S, B and L) in Figure~\ref{fig:PV}. In the setting pair denoted $(A, B)$, $A$ refers to the setting where the stride size is half of the patch size, and $B$ refers to the setting where the stride size is equal to the patch size. As seen in Figure~\ref{fig:PV}, the dip is limited to 1\%, and the increase is significant, reaching approximately 10\% for the EarVN1.0 data set (case: patch = 56, stride = 28).

 The overall trend in our experimental results is evident: configurations with $S = P/2$ yielded superior performance in 44 of 48 experiments, as analyzed in Table-\ref{tab:model_metrics}. This stride setting helped the model acquire fine ear features that span numerous patches, which might be overlooked with non-overlapping patches. These findings underscore the importance of stride setting in obtaining optimal model performance. Hence, in many cases, we observed that the ViT\_T\_p28\_s14\ model is the best-performing model among all configurations and patch-stride settings across all datasets, followed by the ViT\_L\_p16\_s8 model.

\subsection{Dataset based Performance Results}

    \begin{figure}[th]
      \centering
      
      \includegraphics[width=0.5\textwidth]{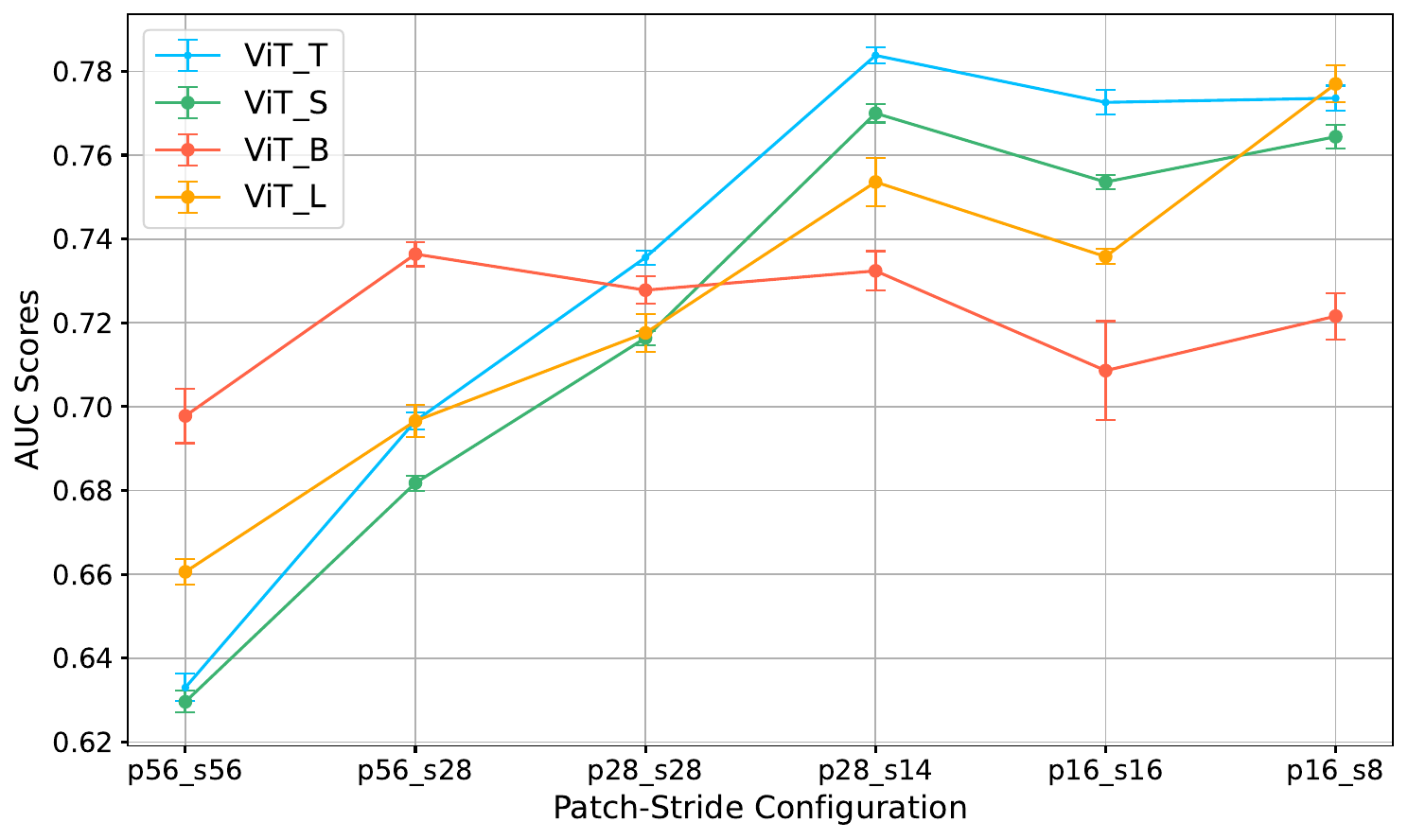} 
      \caption{Comparison of Area Under the Curve (AUC) scores for different patch and stride configurations across all models for the EarVN1.0 dataset.}
       \label{fig:earvn}
       
    \end{figure}

    \begin{figure}[th]
      \centering
      
      \includegraphics[width=0.47\textwidth]{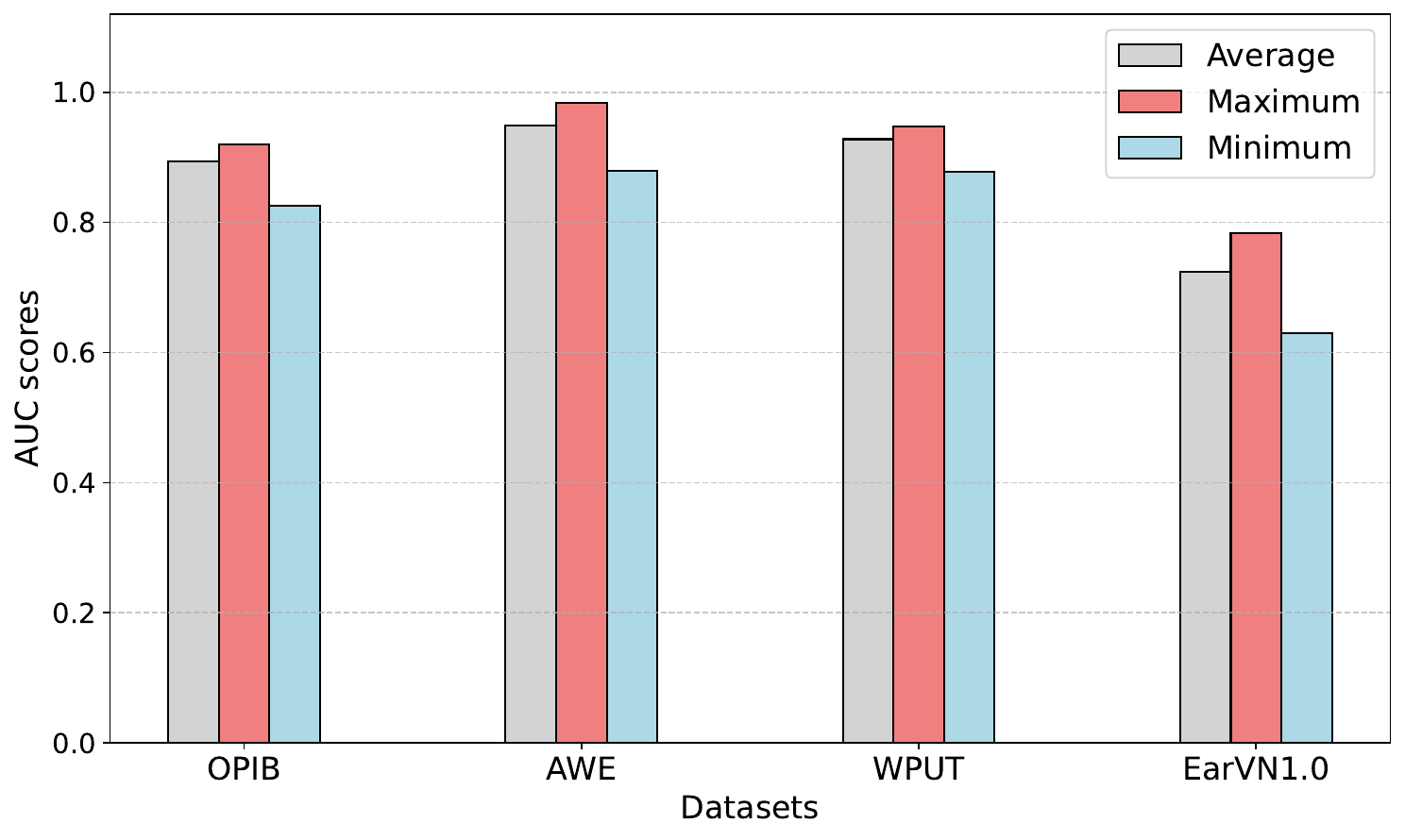} 
      \caption{Maximum, Minimum, and Mean AUC scores for all the datasets across all the ViT models.}
       \label{fig:AUC_min_max_avg}
       
    \end{figure}

The AWE dataset achieved the highest AUC score of 0.9834 ± 0.0011. The OPIB dataset, focused on African subjects with specific pose angle variations, saw a strong performance yielding an AUC score of 0.9196 ± 0.0026. The WPUT dataset, characterized by a balanced gender distribution and high occlusion rate, performed similarly to the AWE dataset, with an AUC score of 0.9472 ± 0.0022. The EarVN1.0 dataset demonstrated lower performance than the AWE, OPIB, and WPUT datasets, with the ViT\_T\_p28\_s14\ achieving an AUC score of 0.7838 ± 0.0019. Despite the low performance, EarVN1.0 showed the highest percentage gain across all datasets, as depicted in Figure~\ref{fig:earvn}, when the half stride setting was used, marking its potential for further improvement. 

We further investigate the overall maximum, minimum, and mean AUC score values for all the datasets across all the ViT models. For the OPIB dataset, we get an average score of 0.8931, with a maximum of 0.9196 and a minimum of 0.8258; for the AWE dataset, we get an average score of 0.9489, a maximum of 0.9834, and a minimum of 0.8796; for the WPUT dataset, we get the average score as 0.9278, maximum as 0.9472, and minimum as 0.8782; and for the EarVN1.0 dataset, we get an average score as 0.7241, maximum as 0.7838, and minimum as 0.6296. For a visual presentation, we also provide a bar chart in Figure ~\ref{fig:AUC_min_max_avg}. 

\section{CONCLUSION}

In this study, we evaluated the performance of the Vision Transformer (ViT) models, ViT-Tiny (ViT-T), ViT-Small (ViT-S), ViT-Base (ViT-B), and ViT-Large (ViT-L)—across diverse datasets using various patch size and stride configurations. Among the models tested, ViT-T consistently outperformed the others in the majority of the cases, achieving the highest AUC scores across three datasets. The configuration ViT\_T\_p28\_s14\ (patch size $P=28$ and stride $S=14$) demonstrated superior performance, particularly on the AWE, WPUT, and EarVN1.0 datasets followed by ViT-L, ViT-S and ViT-B respectively.

The study highlights the critical role of stride configurations in achieving optimal verification performance. Configurations with stride sizes set to half of the patch size (overlapping patches) consistently delivered better results than those with stride sizes equal to the patch size (non-overlapping patches), performing better in 44 of 48 cases. The performance is significant, reaching up to 10\% for the EarVN1.0 dataset (patch size = 56, stride = 28). Smaller patch sizes, such as 16x16, showed mixed results with small models, ViT-S and ViT-T, with a 1\% marginal dip in a few cases. But overall, it is evident that overlapping patches improve model performance. 

Performance also varied significantly across the datasets, reflecting their unique characteristics. The AWE dataset achieved the highest overall AUC score of 0.9834 ± 0.0011. The OPIB dataset, consisting only of African subjects with pose variations, demonstrated strong performance, achieving an AUC of 0.9196 ± 0.0026. Despite its high occlusion rate, the WPUT dataset performed comparably to AWE, with an AUC of 0.9472 ± 0.0022. The EarVN1.0 dataset demonstrated lower AUC scores but showed the highest percentage improvement with half-stride settings, showing possibility for future optimization. Despite the diversity of datasets, our models continuously performed well, demonstrating their durability and flexibility.

Finally, ViT-T, when combined with optimal patch-stride parameters, shows promise as a lightweight, high-performance solution for ear verification applications. These findings open the way for future breakthroughs in biometric applications, particularly in resource-constrained circumstances, by employing efficient configurations of Vision Transformers.


\section{ETHICAL IMPACT STATEMENT}

Ethical considerations are paramount in biometric technology research. While our work advances biometric technology, we acknowledge the need for careful implementation to prevent misuse in surveillance or privacy-invasive scenarios. To mitigate privacy risks, we exclusively utilized anonymized datasets and further anonymized any remaining subject information prior to use in our experiments, ensuring robust protection of individual identities throughout our research process.

In conducting our study, we adhered strictly to ethical research practices. We used the existing datasets that are publicly made available adhering to the guidelines ensuring responsible use of existing resources. No human data was additionally collected for the experiments used in this paper, minimizing privacy risks and potential biases. We also actively worked to identify and mitigate potential algorithmic biases in our models by considering a comprehensive and diverse training set and rigorously testing the models across diverse ear datasets to ensure fair and equitable performance across diverse populations.  

{\small
\bibliographystyle{ieee}

}

\end{document}